\documentclass[10pt,twocolumn,letterpaper]{article}

\usepackage{iccv}
\usepackage{times}
\usepackage{epsfig}
\usepackage{graphicx}
\usepackage{subfigure}
\usepackage{amsmath}
\usepackage{amssymb}
\usepackage{multirow}
\usepackage{bm}

\usepackage[breaklinks=true,bookmarks=false]{hyperref}

\iccvfinalcopy 

\def\iccvPaperID{3848} 

\begin{document}

\title{Design Light-Weight 3D Convolutional Networks for Video Recognition: Temporal Residual, Fully Separable Block, and Fast Algorithm}

\author{Haonan Wang, Jun Lin, Zhongfeng Wang\\
Nanjing University \\
{\tt\small hnwang@smail.nju.edu.cn, \{jlin, zfwang\}@nju.edu.cn}}

\maketitle

\begin{abstract}
    Deep 3-dimensional (3D) Convolutional Network (ConvNet) has shown promising performance on video recognition tasks because of its powerful spatio-temporal information fusion ability. However, the extremely intensive requirements on memory access and computing power prohibit it from being used in resource-constrained scenarios, such as portable and edge devices. So in this paper, we first propose a two-stage fully separable block to significantly compress the model sizes with little accuracy loss. Then a feature enhancement approach named temporal residual gradient is proposed to improve the compressed model performance on video tasks, which provides higher accuracy, faster convergency and better robustness. Moreover, in order to further decrease the computing workload, we propose a hybrid Fast Algorithm to incredibly reduce the computation complexity of convolutions. These methods are effectively combined to design a light-weight and efficient ConvNet for video recognition tasks. Experiments on the popular dataset report $2.3\times$ compression rate, $6.8\times$ workload reduction, and $2\%$ top-1 accuracy gain, over the state-of-the-art SlowFast model, which is already a compact-designed model. The proposed methods also show good adaptability on traditional 3D ConvNet, leading to $5\times$ compact model, $10\times$ less workload, and $3\%$ higher accuracy.
\end{abstract}

\section{Introduction}
\begin{figure}[t]
        \centering
        \includegraphics[height=1.6in, width=3.25in]{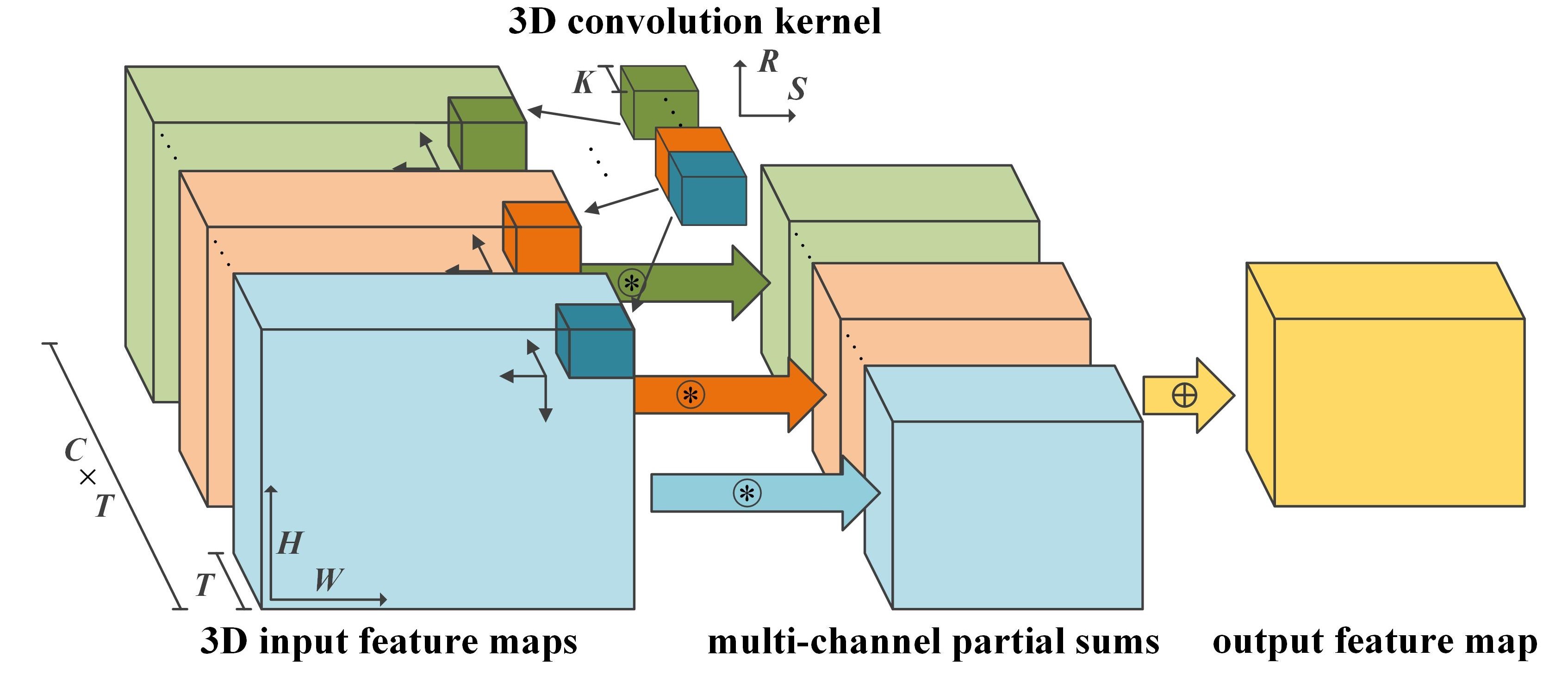}\\
        \caption{\textbf{Illustration of a convolutional layer}, in which multi-channel input feature maps are convolved with one convolution kernel along spatial and temporal dimensions, yielding one output feature map.}\label{fig:3Dconv}
    \end{figure}

    Deep Convolutional Network (ConvNet) has demonstrated remarkable performance on visual tasks with still image, such as document recognition~\cite{lecun1998gradient}, object detection~\cite{girshick2015fast}, and semantic segmentation~\cite{chen2014semantic}. It employs 2-dimensional (2D) convolutional layers as basic blocks to effectively learn spatial features. But the direct transplantation of 2D ConvNets are not suitable for video tasks, such as action recognition~\cite{baccouche2010action,baccouche2011sequential}, because of the lack of temporal information fusion. Inspired by the topological structure of 2D ConvNets, many recent works expand them to 3-dimensional (3D) ones---an extra temporal dimension is added---to handle with the end-to-end video tasks. As shown in Fig.~\ref{fig:3Dconv}, it illustrates a 3D convolutional layer with multiple input channels and only one output channels. It can be seen that the 3D convolution kernel jointly modeling the spatio-temporal information by drifting along both pixel and frame orientations.

    In the work of C3D~\cite{tran2015learning}, it explored convolution kernels with size of 3$\times$3$\times$3, which is empirically proved to work best for all layers among limited architecture exploration. However, models that simply stacking 3D convolutional blocks layer by layer cannot reach the similar accuracy with respect to their 2D counterparts for image classification tasks~\cite{he2016deep}, even though the datasets to be trained on for the former are far small than that of the latter (ImageNet). Hence, many works start to rethink of the effectiveness of 3D ConvNet. In~\cite{zhou2018mict}, it is argued that the high model complexity attributed to massive 3D convolution kernels hinders the model itself from converging to an optimal local minimum. On the other hand, the incredible large model size of 3D ConvNets brings considerable memory access power consumption, and the computation complexity grows exponentially with regard to the model size, which will require more computing power and more training time. Although video recognition applications under resource-restrict scenarios are on a more urgent demand, the above two factors make it more difficult to train a deeper 3D-kernel-stacked ConvNet, in addition, show few possibilities to deploy such models on portable or edge devices. As a result, it is highly desired that a compact and computational efficient 3D ConvNet, which only consumes limited computing resources and reach smaller model size.

    In this paper, we propose a two-stage fully separable block (FSB) as a basic block to construct a compact 3D ConvNet. It can be employed to significantly compress the size of naive 3D models and that of already compact-designed ones \eg SlowFast network, while only resulting in slight accuracy loss on various typical video recognition tasks.  Nevertheless the FSB can considerably shrink the model size, the massive operations brought by the computationally intensive convolutions is still unacceptable for the deployment of video-tasks-aimed models on budget-restricted platforms. Since the convolution operations dominate the overall computation, we propose a hybrid Fast Algorithm (hFA) to significantly reduce the computation complexity. Moreover, a feature enhancement approach named temporal residual gradient is proposed to improve the model performance on video tasks, providing higher accuracy, faster convergency and better robustness. Based on an effective coupling of these approaches, we design a light-weight and efficient network for video tasks. Experimental results show that the proposed model achieves an overwhelming performance of $2.3\times$ compression rate, $6.8\times$ workload reduction, and $2\%$ top-1 accuracy gain, over the state-of-the-art compact-designed SlowFast network~\cite{feichtenhofer2018slowfast}. We also evaluate our methods on the conventional C3D model~\cite{tran2015learning} to verify their adaptability, leading to $5\times$ compression rate, $10\times$ workload decreasing, and $3\%$ accuracy gain.


\section{Related Work}

\textbf{3D ConvNets and Compact approaches.} Many recent works~\cite{ji20133d,varol2018long} proposed 3D ConvNet models for human action recognition. Based on a limited exploration of the architecture design space, \emph{Tran} \etal~\cite{tran2015learning} proposed a more powerful network, in which it claimed that the convolution kernels with size of $3\times 3\times 3$ are empirically optimal for all layers. Hence, many follow-up works preserved the backbone of the $3\times 3\times 3$ kernels with peripheral modifications [I3D]. However, the exponential increasing scale on model complexity made it harder to train a deeper 3D ConvNets, because the optimal converge is less possible to be found. Hence, some recent works started to rethink the necessity of the employment of such computationally intensive operation block. \emph{Zhou} \etal ~\cite{zhou2018mict} argued that learning spatio-temporal information by simply stacking 3D convolutions layer by layer hindered the SGD optimizer from converging to the global optimum, since the parameter searching space exponentially increased as the model went deeper. So they cascaded 2D convolutions with 3D ones as compact structures to reduce the massive parameters. \emph{Xie} \etal~\cite{xie2018rethinking} found that separately learning the spatial and temporal features leaded to a better result by replacing the 3D convolutions with 2D/1D cascaded ones, even though the complexity of the latter was lower. He \etal [slowfast] also introduced this bottleneck structure to their recent presented work, which is based on the two-stream structure~\cite{feichtenhofer2016spatiotemporal,feichtenhofer2016convolutional,simonyan2014two}, achieving the state-of-the-art performance. It leveraged two path to jointly cope with the video tasks, in which the fast path was designed light-weight to learn movements, while the slow one mainly used for recognizing spatial instances.

\textbf{Fast Algorithm for low-complexity ConvNets.} The application of Fast Algorithms in ConvNets was first proposed by~\cite{mathieu2014fast,vasilache2014fast}. Then Lavin \etal~\cite{lavin2016fast} applied another type of fast approach called Winograd Algorithm (WinoA)~\cite{winograd1980arithmetic} for 2D convolutions, leading to a great compression rate on the convolutional layers. It have been integrated into the cudnn library~\cite{chetlur2014cudnn} as a build-in method, which proved the superior of the WinoA in convolution implementations. Recently the 3D WinoA was also proposed to decrease the computation complexity of 3D ConvNet~\cite{shen2018towards}, although it could only be used for 3D kernels with same scales on temporal and spatial orientations.


\section{Design Compact and Efficient Networks}

In the following sections, we will give a detailed illustration of the proposed FSB structure and its heuristic motivation, as well as how it significantly compresses the state-of-the-art 3D ConvNets. Then the theoretical derivations of the Fast Algorithms are introduced . Moreover, it is presented that how these algorithms are combined to further reduce the computation workload of 3D models to an extremely low level.  Due to the main idea of this section is to show the compression ability of the proposed method, more evaluations on the robustness and the performance of our approaches will be illustrated in Section~\ref{sec:experiment}.

\subsection{Naive 3D Convolution}\label{sec:naive3d}
For the simplicity of description, in this section, we view the convolution kernels with multiple input and output channels as a 5-dimensional tensor. As shown in Fig.~\ref{fig:3Dconv}, in a naive 3D convolutional layer, an extra dimension is expanded along the temporal dimension compared to the 2D ones. It is illustrated that a convolution kernel with only one output channel learns the spatio-temporal features by convolving with the multi-channel input feature maps (ifmaps) along the vertical, horizontal and temporal directions, where $\{T,H,W\}$ and $\{K,R,S\}$ denote the temporal duration and spatial sizes of the ifmaps and kernels, respectively, and $C$ represents the number of input channels. These convolution operations in different input channels are performed independently, then the convolved partial sums need to be accumulated throughout all channels into one output channel. Normally multiple kernels are utilized to yield certain a number of output channels, of which the number is notated as $N$. And the zero-padding strategy is usually used to maintain the size unchanged between clips in the ifmaps and the ofmaps. The mathematical form of the 3D convolution is as follows:
\begin{align}\label{eq:3Dconv}
  Y_{i,k}&=D_{i}*G_{k}, \\
  y_{i,k,t,x,y}&=\sum\limits^C_{c=1}\sum\limits^T_{s=1}\sum\limits^R_{v=1}\sum\limits^S_{u=1}d_{i,c,t+s,x+u,y+v}g_{k,c,s,u,v},
\end{align}
where we denote the $i$-th ifmap and the $k$-th kernel as $D_i$ and $G_k$, respectively. It can be found that the computation complexity exponentially increases as the model size raising.

\subsection{Temporal Residual Gradient Module}
\begin{figure}
  \centering
  \includegraphics[height=1.6in, width=3.25in]{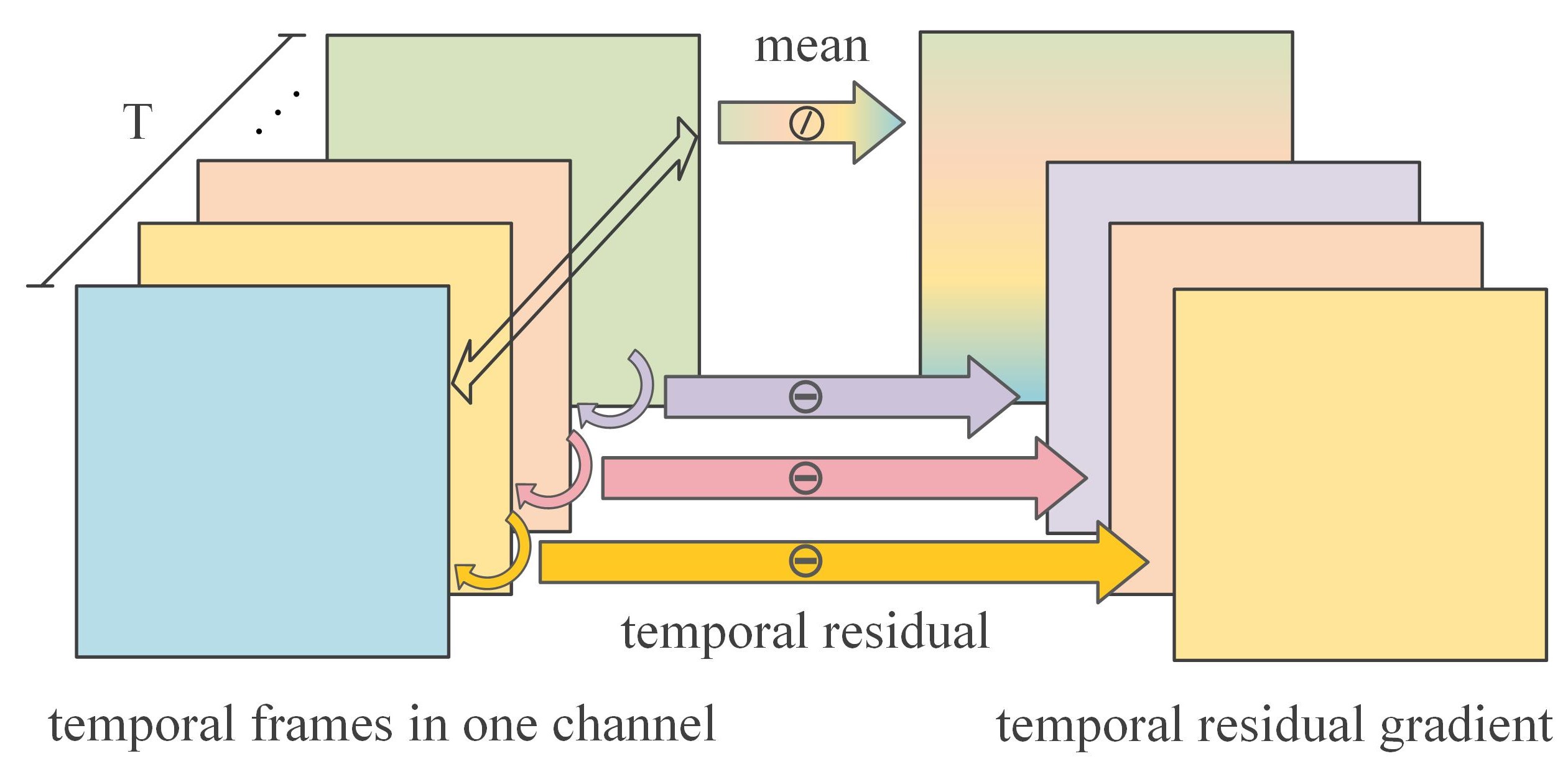}\\
  \caption{TRG structure}\label{fig:trg}
\end{figure}

We propose a heuristic method called temporal residual gradient (TRG) that offers the model more robustness and faster convergency based on a basic priori acknowledge---the gradient between adjacent image frames indicates the motions of the objects. As shown in Fig.~\ref{fig:trg}, the TRG computes the residual value between each two frame on temporal orientation to represent the movement in time domain, yielding $T-1$ frames of gradient features. And in order to preserve the statistical distributions of the original features, mean values throughout the temporal dimension are calculated and cascaded with the residual gradient features, getting a output tensor with total $T$ frames. Similar to the Histogram of Oriented Gradient (HOG)~\cite{llorca2013vehicle,pang2011efficient} features in 2D image object detection tasks, which forms the features by computing the spatial gradients of local field, the TRG module can extract the active motion features with an extra frame of the approximate description of the action scenario. It relieves the burden of jointly learning spatio-temporal features with 3D kernels and shrinks the searching space of the optimization operator, offering the model higher performance and faster convergency. More detailed discussions on the TRG method are demonstrated in Section~\ref{sec:experiment}. On the other hand, the $T-1$ frames output from TRG are performed subtractions, and the last one with the mean value can be calculated by addition and bit shifting operations, considering the number of frames it divided by is often the power of $2$. So the TRG is efficient in hardware implementation.

\subsection{Fully Separable Block}

\begin{figure*}[t]
\begin{center}
    \includegraphics[height=2.3in, width=6in]{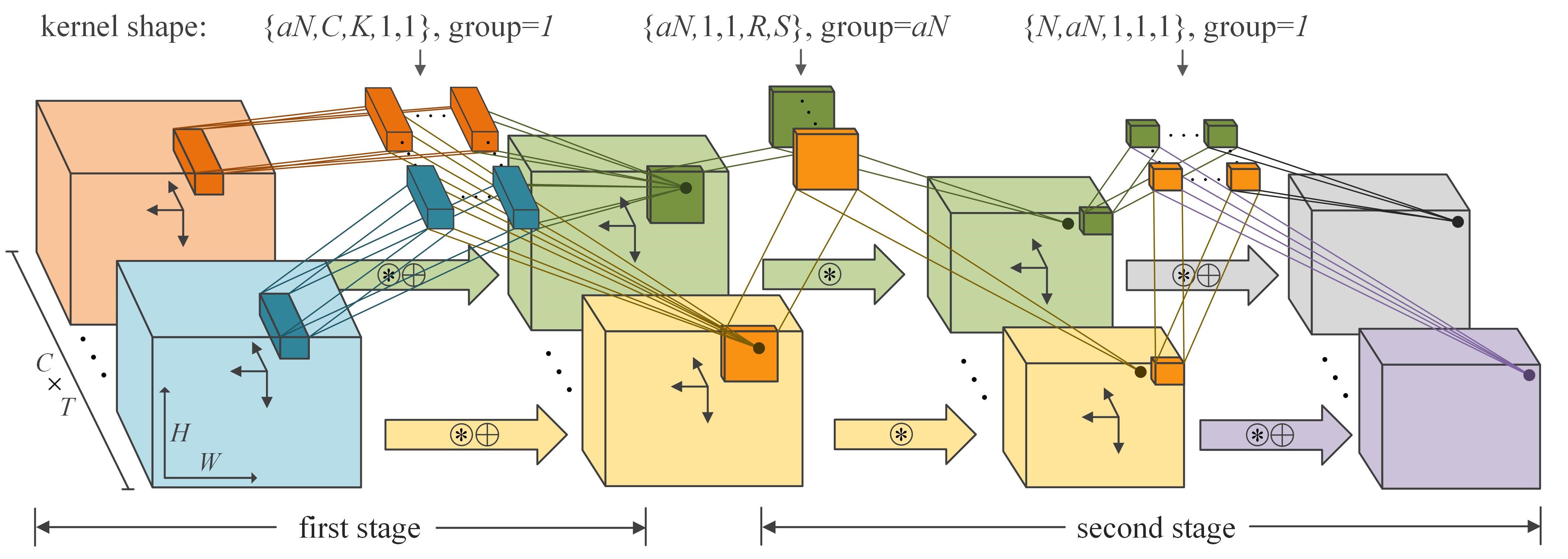}\\
\end{center}
   \caption{\textbf{Illustration of the FSB structure}, where no zero-padding is used for better understanding of the dataflow.}
\label{fig:FSB}
\end{figure*}

According to Section~\ref{sec:naive3d}, the computation complexity and model size of the naive 3D ConvNet are totally unacceptable for deployment under resource-constrain applications. So we propose a compact but effective two-stage fully separable block structure, which can significantly slim the ConvNets yet making no harm to the accuracy. As illustrated in Fig.~\ref{fig:FSB}, the FSB leverages several separated small steps to gradually fuse the fine-grained spatio-temporal information. At the first stage of the FSB, a temporal bottleneck structure is hired. In this phase, the 3D convolution kernel is separated into two components, of which the first focuses on merge pixels along the temporal orientation and the second employs 2D kernels to learn spatial features of different frames. As for the second stage, we further split the channels apart, each of which convolved with a 2D kernel without inter-channel communication, because we then leave them to the last step, where $1\times 1\times 1$ kernels are used to jointly accomplish both spatial and temporal information fusions, as well as the output channel reshaping.


In addition to a heuristic description of the topological structure, we then give a detailed analysis about the considerable compression rate achieved by replacing the naive 3D convolutional layer with the proposed FSB. Considering a convolutional layer that receives $C$ input clips with size of $\{T,H,W\}$ and needs to yield ofmaps consisted of $N$ channels, it rules the shape of kernels as $\{N, C, K, R, S\}$, leading to memory requirement of $N\times C\times K\times R\times S$. As for the FSB, in the first stage, a bottleneck structure---kernel with size of $\{\alpha N, C, K, 1, 1\}$---is employed to focus on the temporal fusion, where $\alpha$ denotes the intermediate expansion rate. Next, a $\{\alpha N, 1, 1, R, S\}$ kernel with $\alpha N$ groups is hired to learn intra-channel spatial features, which means each group only need to be convolved with one corresponding channel. At last, a 3D point-wise convolution is performed via a $\{N,\alpha N, 1, 1, 1\}$ to finally combine the spatio-temporal information separably located in different channels, in addition, to reshape the output tensor to fit the input size constrain of the next layer. As a result, a convolutional layer that leveraging the proposed FSB merely consumes total memory footprint of $\alpha NCK+\alpha NRS+\alpha N^2$. Considering a typical case of $\{N,C,K,R,S\}=\{64,64,3,3,3\}$, and we set the $\alpha$ to 1 in this paper, the FSB can compress a convolutional layer by $6.5\times$.

As the stacked 3D layers in a naive 3D ConvNet like C3D~\cite{tran2015learning} can be easily replaced, the FSB-based model can achieve great compression rate. Moreover, even for an already compactly designed network, \eg the state-of-the-art SlowFast network~\cite{feichtenhofer2018slowfast}, the compression performance can still be improved by replacing the 3D residual blocks with our FSBs. Detailed compression of different 3D ConvNets on model size are listed in ~\ref{tab:accuracy}.


\subsection{Efficient Hybrid Fast Algorithm}

Even though the proposed FSB leads to considerably small model size, the overall computation complexity of these models still remains at a high level due to its reliability not only on the sizes of weights but also that of activations, the shape of which is determined by the fixed topological architecture and unlikely to be compressed. Hence, a hybrid Fast Algorithm is proposed to directly decrease the computation strength of convolutions in the FSB, as we refer to the number of multiplication operations as an evaluation of the model computation complexity, since the multiplications far more implementation-expensive than additions and they dominate the computational time and power of the deployment of a ConvNet~\cite{wang2016efficient}. In the following subsections the Fast Algorithms are theoretically introduced, and we will propose a hybrid dataflow, which couples these methods, to incredibly reduce the computation complexity of a 3D convolutional layer.

\subsubsection{Fast Algorithm} Fast Algorithm includes a cluster of computation strength reduction methods, which are initially introduced and exploited in the signal processing field. Several typical algorithms were applied to improve the convolution computation efficiency, such as the Fast Fourier Transform (FFT)~\cite{mathieu2014fast}, the Fast FIR~\cite{mou1987fast,wang2018efficient}, and the WinoA~\cite{lavin2016fast}, so they could be implemented in the ConvNets naturally. Another work named Strassen Algorithm was introduced to reduce the multiplication operations of the matrix multiplication. Considering the regularity and adaptability of the dataflow, We propose the hybrid Fast Algorithm based on the WinoA in this paper.

\textbf{1D WinoA}. The WinoA is first introduce to reduce the FIR filter complexity. For a $r$-tap FIR aimed at $m$ outputs, which can be viewed as a 1-dimensional (1D) convolution,  The traditional convolution can be written in matrix form as
\begin{equation}\label{eq:1dconv}
  Y=g*d,
\end{equation}
where the notation $*$ represents convolution operation. The 1D WinoA, which is denoted as $F(m,r)$, re-express the Eq.~\ref{eq:1dconv} as follows,

\begin{equation}\label{eq:1dwino}
  Y=A^T[(Gg)\odot(B^Td)],
\end{equation}
where $\odot$ indicates the element-wise matrix multiplication (EWMM), and the $A$, $G$, and $B$ denote the transform matrices. What should be mentioned is that these transform matrix are only consisted of 0, 1, and values of powers of 2, which can be implemented using low-cost bit-shift operations, so multiplications are only performed in the EWMM operations. As the $g$ and $d$ are both transformed into the shape of $Y$ by $G$ and $B$, separately, the number of multiplications equals to the order of matrix $Gg$ (\ie $m+r-1$). Hence, for a 1D convolution, the computation complexity can be reduced from $m\times r$ to $m+r-1$. One can refer to~\cite{lavin2016fast} for a more detailed theoretical derivation. In the first stage of the FSB, all the bar stretched along the temporal direction within a channel clip can employ the 1D WinoA.

\textbf{2D WinoA}, denoted as $F(m\times m,r\times r)$, is performed on a local 2D block, which convolves a $(r\times r)$ kernel $g$ with a input tile $d$, yielding an output tile $Y$ with size of $(m\times m)$. Hence, the input $d$ should be a $(m+r-1)\times(m+r-1)$ tile. By applying the 2D WinoA, the 2D convolution operation can be re-expressed as

\begin{equation}\label{eq:2dwino}
  Y = A^{T}[(GgG^{T})\odot(B^{T}dB)]A,
\end{equation}
so the 2D WinoA thus only requires $(m+r-1)^2$ multiplications, while the traditional convolution needs $m^{2}\times r^{2}$. In the second stage of the FSB, the 2D WinoA is employed to compress the fully separated 2D kernels by iteratively convolving them with input tiles in each temporal frame with $(r-1,r-1)$ strides.

\textbf{3D WinoA.} 3D WinoA is recently proposed in~\cite{shen2018towards}. Similarly, it is represented as $F(m\times m\times m, r\times r\times r)$. The theoretical representation of the 3D WinoA can be heuristically derived by expanding the Eq.~\ref{eq:2dwino} into 3 dimensions as shown in Eq.~\ref{eq:3dwino},

\begin{equation}\label{eq:3dwino}
  Y = \{A^{T}[(GgG^{T})^{R}G^{T}\odot(B^{T}dB)^{R}B]\}^{R}A,
\end{equation}
where operator $R$ means transposition between the 2-$nd$ and 3-$rd$ dimensions. This method only slightly improves the computation complexity reduction compared with its 2D counterpart ($<15\%$), but it will incredibly increase the transform operations. Otherwise, it can only be used by 3D kernels, which is lack of adaptability.



\subsubsection{hybrid Fast Algorithm}
The 3D WinoA is introduced to make a comparison of our proposed compression method with the state-of-the-art 3D WinoA. Since a 3D convolution kernel is decomposed into small ones with different in the FSB, we need to exploit hybrid types of the Fast Algorithms to efficiently cope with all kinds of decoupled kernels within the fine-grained stages of the FSB. At the first stage, a temporal kernel with scale of $K\times1\times1$ can utilize the $F(m_1,K)$, so it results in a complexity reduction by $\frac{m_{1}K-m_{1}-K-1}{m_{1}K}$ times. As for the second stage, one can apply the $F(m_2\times m_2, R\times R)$ to the $1\times R\times R$ separated kernels to reach $\frac{m_2^2 R^2}{(m_2+R-1)^2}$ times of computation compression. So the hFA can be represented as $F(m_1,K)\otimes F(m_2\times m_2, R\times R)$, where $\otimes$ denotes concatenation. We evaluate the hFA in Table.~\ref{tab:hfa}, in which it can be concluded that this dedicated algorithm is more adaptive for mainstream models and more superior in the computation complexity reduction.

\begin{table}[h]
\begin{center}
\begin{tabular}{|l|c|c|}
\hline
\multirow{2}{*}{Method} & \multicolumn{2}{c|}{Number of multiplications}\\
\cline{2-3}
 & C3D~\cite{tran2015learning}  & SlowFast~\cite{feichtenhofer2018slowfast}\\
\hline\hline
Baseline & 199.9G & 145.5G \\
3D Winograd [] & 25.0G & - \\
FSB-only & 27.6G & 68.7G\\
FSB+hFA & \textbf{18.1G} & \textbf{40.7G}\\
\hline
\end{tabular}
\end{center}
\caption{Computation complexity of different models. Evaluations are performed on UCF-101 with input size of $64\times112\times112$.}\label{tab:hfa}
\end{table}

\section{Experiments}\label{sec:experiment}
In this section, we first give a brief introduction about the evaluation datasets, and then some experimental settings are discussed. A comprehensive design space exploration is performed to decide the optimal structure of the FSB and to find out the robust strategy of its combination with the mainstream models. We draw a conclusion from the experimental results that with use of the proposed FSB and hFA methods, the 3D ConvNet models can achieve the similar performance on complicated video tasks, while it only costs incredibly low memory footprint and computation complexity, compared with their original editions.

\subsection{Evaluation Settings}
We evaluate the proposed methods on two representative ConvNet prototypes, which are the C3D, a naive 3D-kernel-stacked network, and the SlowFast network, a compact-designed model. In the experiments, all models are training on the widely-used benchmark UCF-101~\cite{soomro2012ucf101} from scratch without any pre-train. And the optimization phases all use SGD optimizer with momentum on signal GPU for fair comparisons. The UCF-101 is one of the most popular datasets among the video recognition tasks, which contains 13,320 labeled action videos in total with classification of 101 categories. It provide three split lists for dividing the dataset into train/test subsets. In order to evaluate the representative capacity of the proposed model and the baseline, we increase the training and validation sets by ensembling two subsets.


\subsection{Design Space Exploration on SlowFast}
Our work mainly draw inspirations from the state-of-the-art SlowFast network, so we will analysis the difference in detail and explain why the FSB can further improve the performance. Moreover, several variants of the FSB, based on the SlowFast network, are explored to find the optimal topological structure.

\begin{figure*}[t]
    \centering
    \subfigure[training loss]{
    \begin{minipage}[t]{0.33\linewidth}
        \centering
        \includegraphics[width=2.3in]{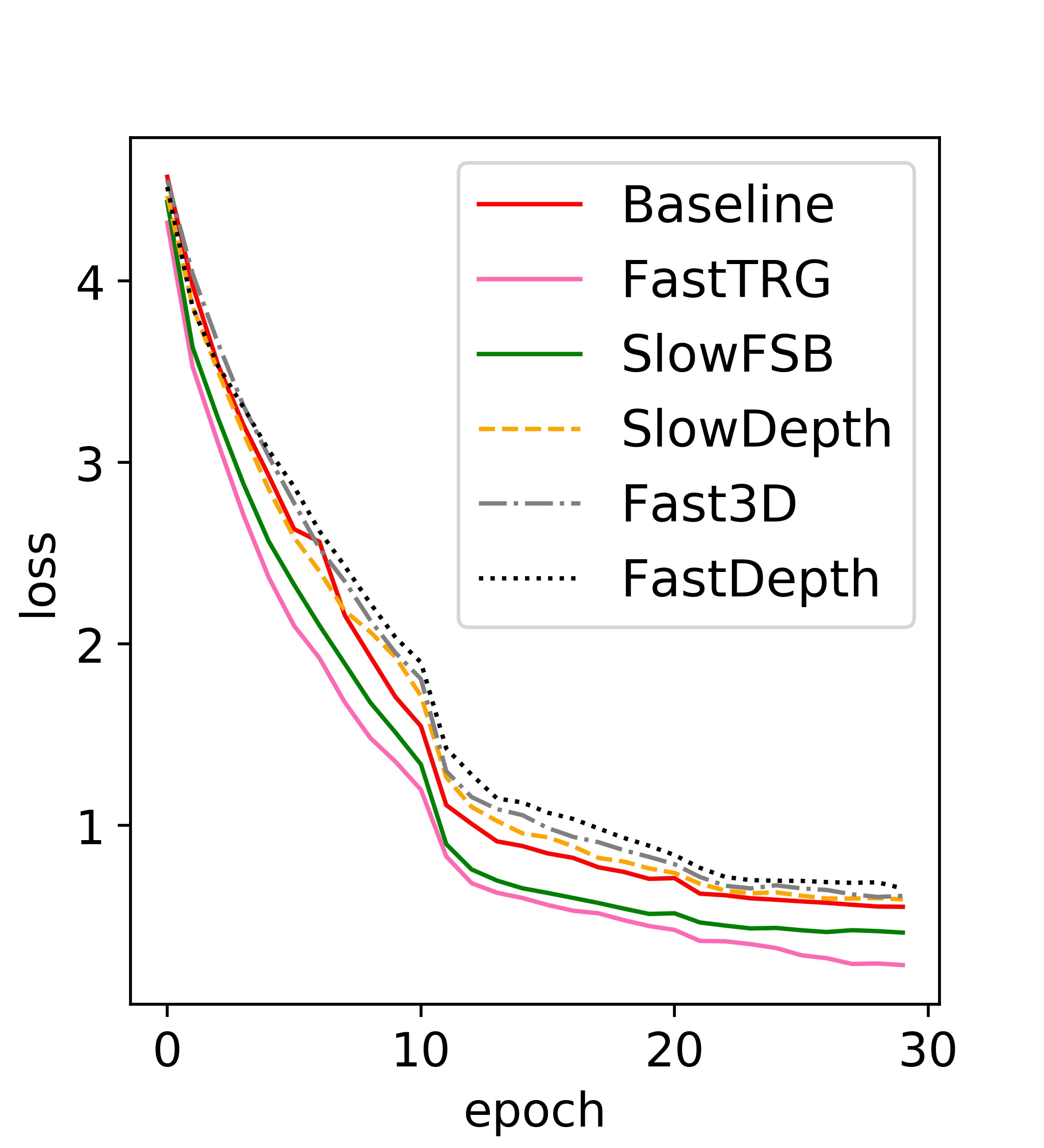}\label{fig:trainloss}
    \end{minipage}%
    }%
    \subfigure[training accuracy - top1]{
        \begin{minipage}[t]{0.33\linewidth}
        \centering
        \includegraphics[width=2.3in]{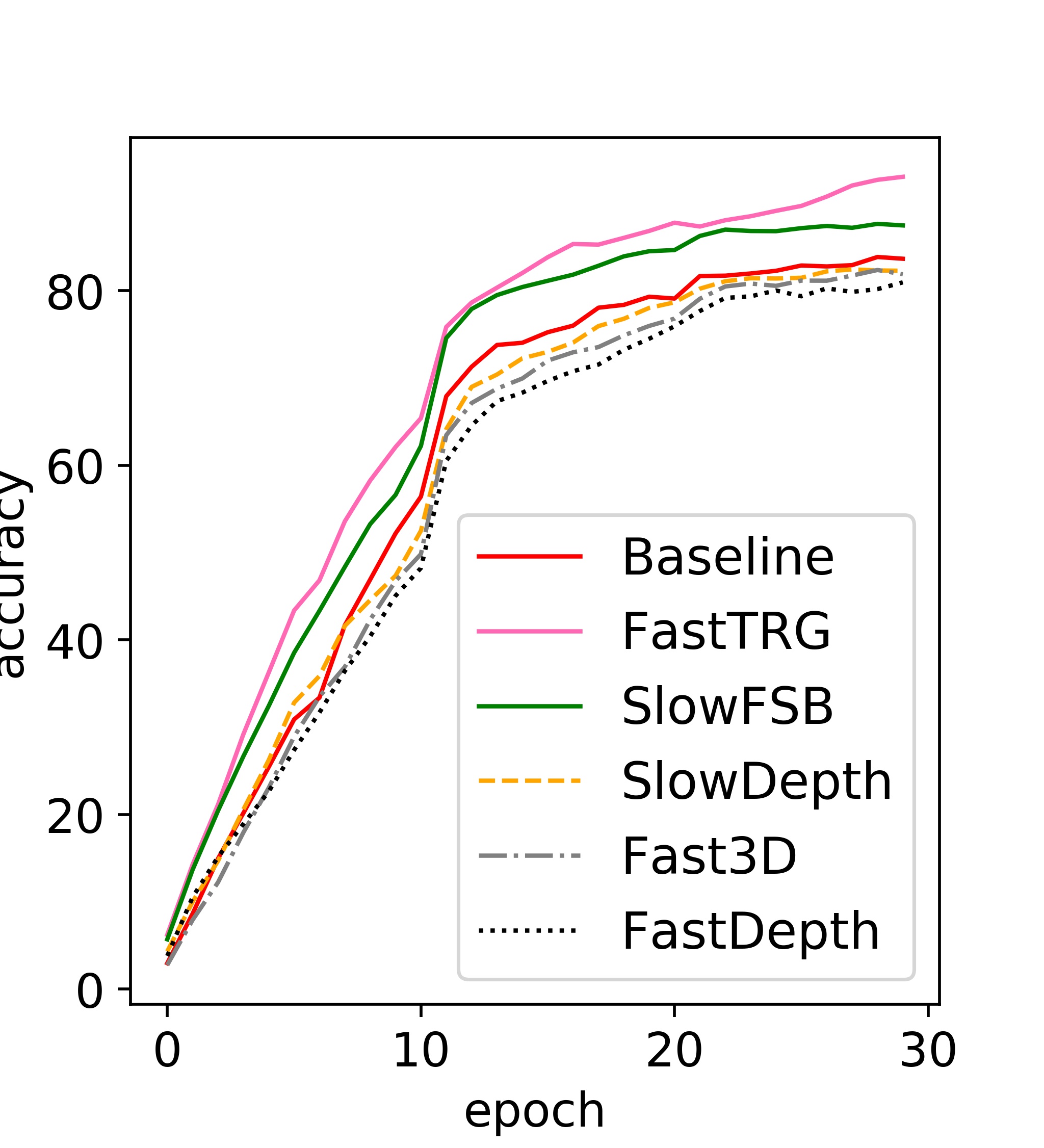}\label{fig:traintop1}
    \end{minipage}%
    }%
    \subfigure[training accuracy - top5]{
        \begin{minipage}[t]{0.33\linewidth}
        \centering
        \includegraphics[width=2.3in]{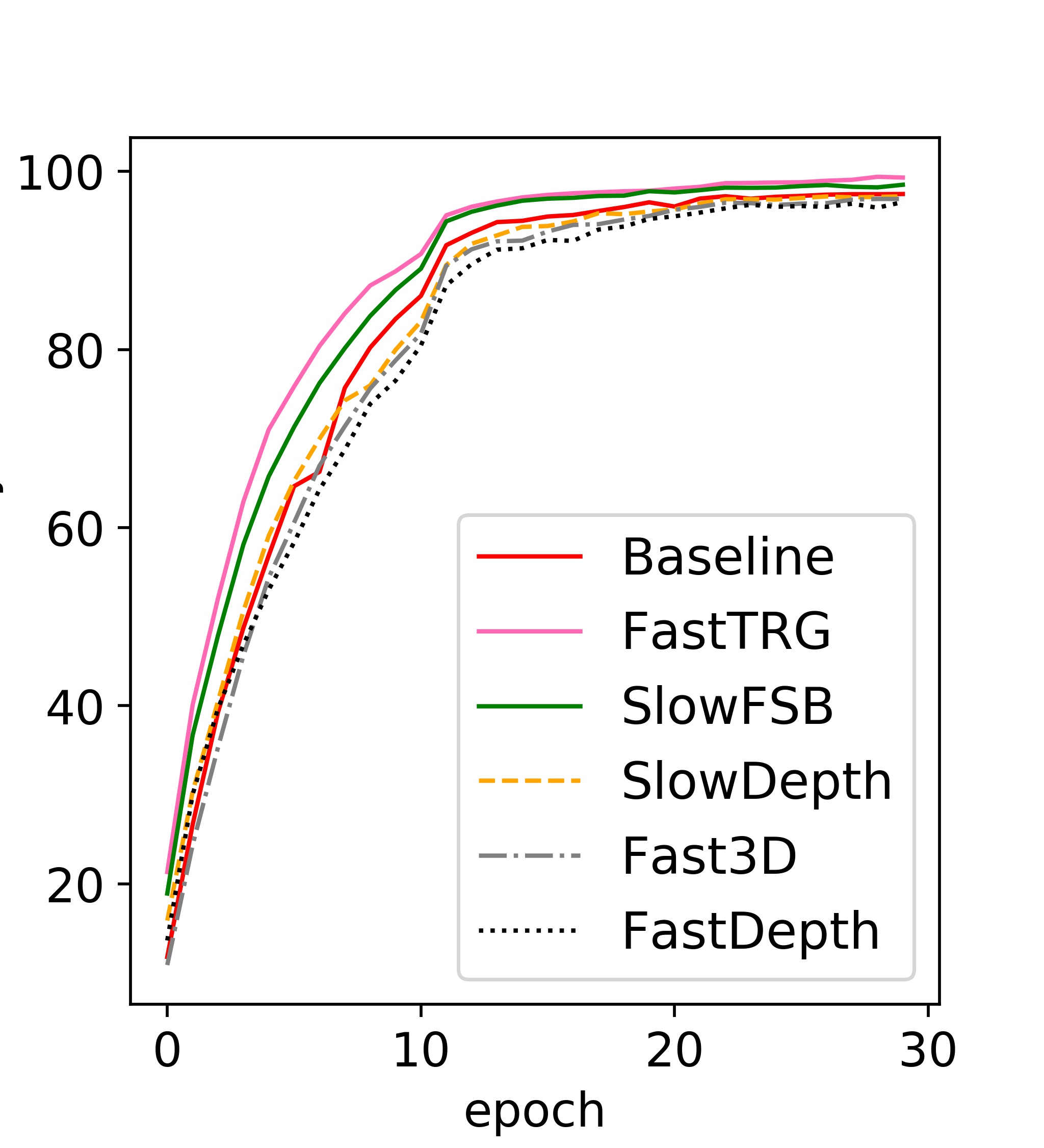}\label{fig:traintop5}
    \end{minipage}%
    }%

    \subfigure[validation loss]{
    \begin{minipage}[t]{0.33\linewidth}
        \centering
        \includegraphics[width=2.3in]{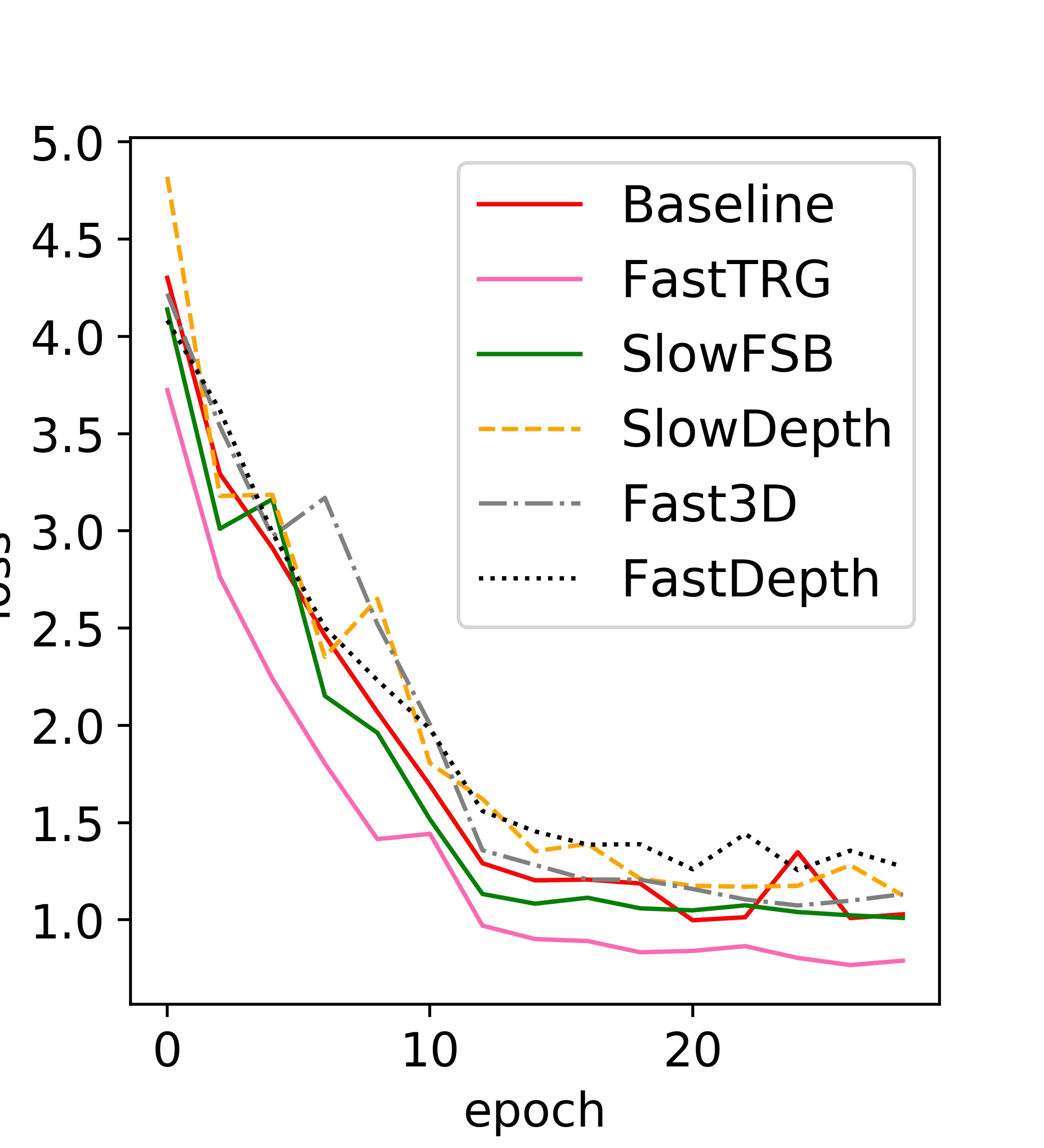}\label{valloss}
    \end{minipage}
    }%
    \subfigure[validation accuracy - top1]{
        \begin{minipage}[t]{0.33\linewidth}
        \centering
        \includegraphics[width=2.3in]{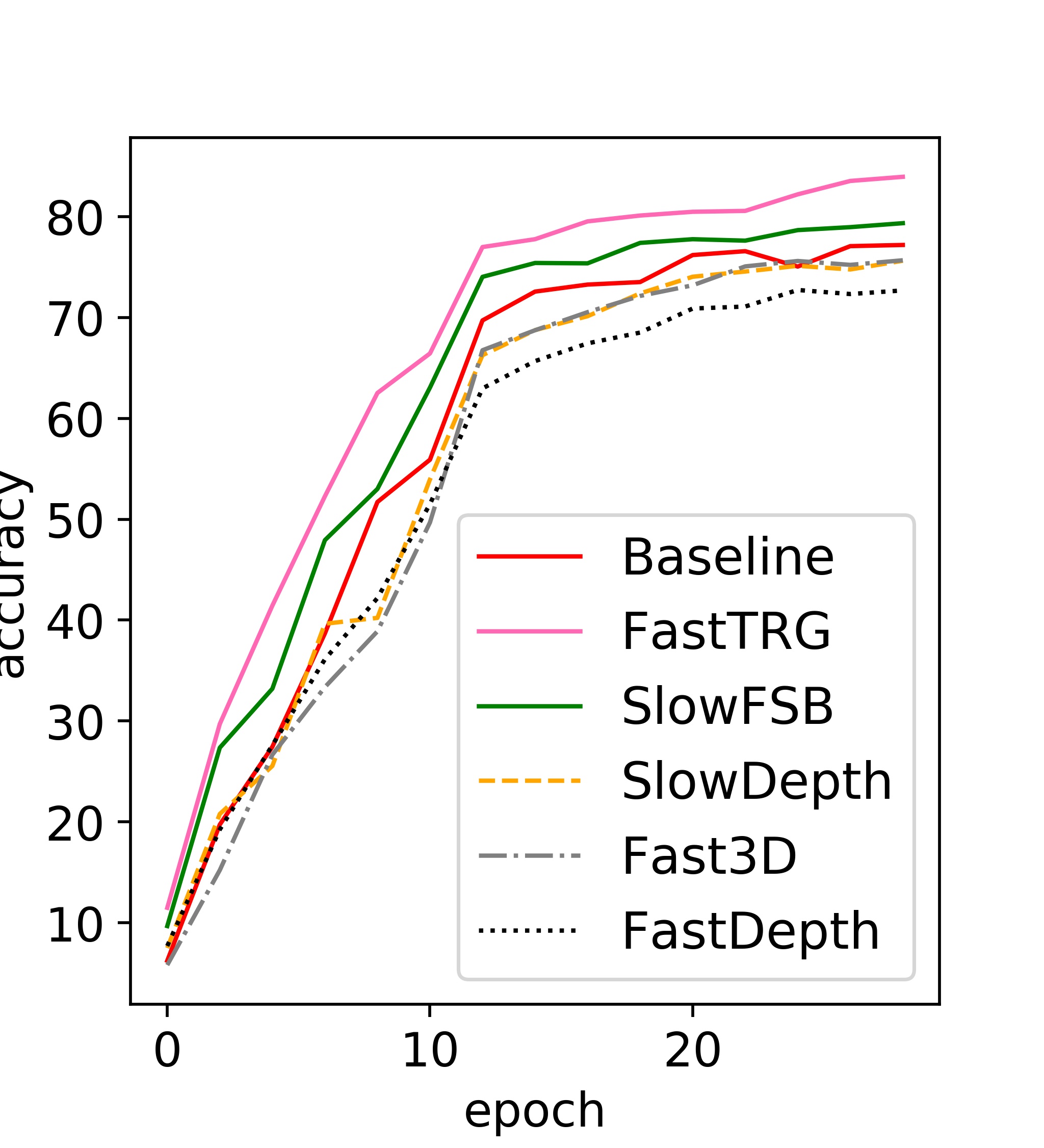}\label{fig:valtop1}
    \end{minipage}
    }%
    \subfigure[validation accuracy - top5]{
        \begin{minipage}[t]{0.33\linewidth}
        \centering
        \includegraphics[width=2.3in]{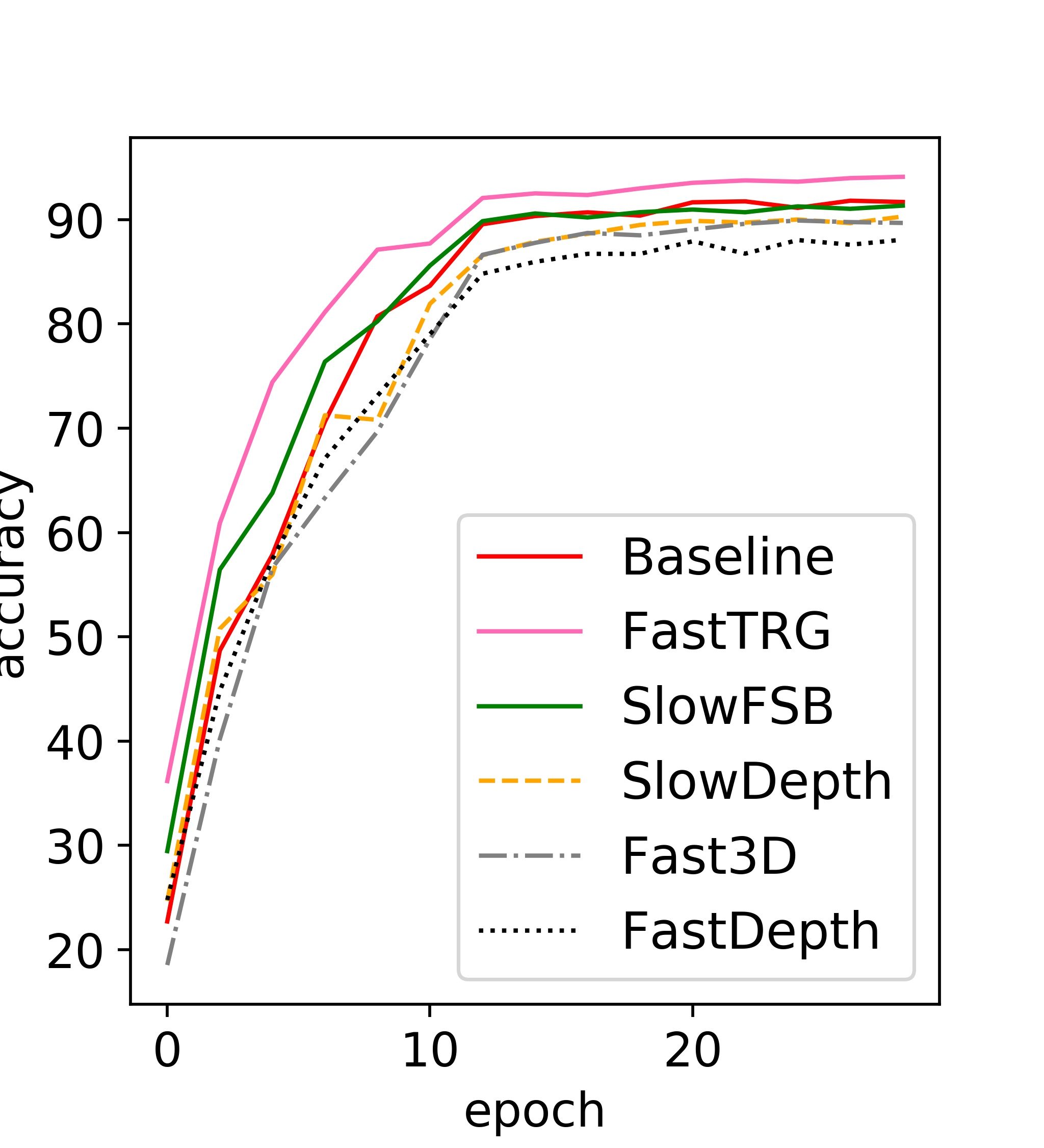}\label{fig:valtop5}
    \end{minipage}
    }%

\centering
\caption{\textbf{Design space exploration.} It is strongly recommended to be viewed in color.}\label{fig:design}
\end{figure*}

The SlowFast network separates a traditional $3\times3\times3$ kernels into a 3-step residual bottleneck structure, which can fuse spatio-temporal information step by step, leading to a more accurate learning process and a more precisely route to the optimal minimum, while taking the advantage of low memory consumption of the bottleneck structure. Following this tendency, we believe that further separating the channel merging process may result in a more fine-grained structure suitable for better convergency and, meanwhile, achieve a more compact model. According to this priori hypothesis, we explore the design space of the FSB structure for an optimal performance and then evaluate these variants on UCF-101 dataset, as shown in Fig.~\ref{fig:design}.

\begin{table*}[t]
\begin{center}
\begin{tabular}{l||c|c|c|c|c|c|c|c}
  \hline\hline
  \multirow{2}{*}{Model} & \multicolumn{4}{c|}{training} & \multicolumn{4}{c}{validation} \\
  \cline{2-9}
  & top-1 acc & top-1 gain & top-5 acc & top-5 gain & top-1 acc & top-1 gain & top-5 acc & top-5 gain \\
  \hline\hline
  Baseline~\cite{feichtenhofer2018slowfast} & 84.1\% & - & 97.8\% & - & 77.6\% & - & 91.8\% & - \\
  \hline
  FastTRG & 92.7\% & \textbf{+8.6\%} & 99.4\% & \textbf{+1.4\%} & 83.9\% & \textbf{+6.3\%} & 94.1\% & \textbf{+2.3\%} \\
  \hline
  SlowFSB & 87.7\% & +3.6\% & 98.5\% & +0.7\% & 79.4\% & +1.8\% & 91.3\% & -0.5\% \\
  \hline
  SlowDepth & 83.4\% & -0.7\% & 97.3\% & -0.5\% & 75.8\% & -1.8\% & 90.4\% & -1.4\% \\
  \hline
  FastDepth & 81.7\% & -2.4\% & 96.7\% & -1.1\% & 73.5\% & -4.1\% & 88.2\% & -3.6\% \\
  \hline
  Fast3D & 83.4\% & -0.7\% & 97.3\% & -0.5\% & 76.4\% & -1.2\% & 90.0\% & -1.8\%\\
  \hline
  \hline
\end{tabular}
\end{center}
\caption{Model Performance on variants.}\label{tab:accloss}
\end{table*}


\textbf{SlowDepth for compression with slight accuracy loss.} Considering that the SlowFast network is based on the two-stream method, in which one light-weight path operates at a high frame rate, which means to capture the temporal features (\ie motions), and another path works at a low frame rate to model the spatial information. We suppose that the slow path mainly play a role of capturing features from still images and providing the object instances information to support the final action recognition, but the slow path dominates the overall memory footprint of the network ($\approx98\%$). According to the general perspective of many works that focus on model compression of 2D ConvNets~\cite{iandola2016squeezenet,wang2018sgad}, we believe that there will be considerable redundancy in the slow path, so a channel bottleneck structure are introduced, which is composed of a depth-wise $3\times3$ kernel and a point-wise $1\times1$ one, to replace the $1\times3\times3$ kernel for further reduction of the model size. This variant is denoted as \textbf{\emph{SlowDepth}} in Fig.~\ref{fig:design}. It can be conclude that this method slightly hurt the final model accuracy of around 1\% (training(top1/top5): from $84.1\%/97.8\%$ to $83.4\%/97.3\%$; validation(top1/top5): from $77.6\%/91.8\%$ to $75.8\%/90.4\%$), while it is achieved that a 1.9$\times$ compression rate in model size over the original SlowFast network, which is also trained on UCF-101 as a baseline. We have compared six variants with the baseline as listed in Table.~\ref{tab:accloss}.

\textbf{SlowFSB for better accuracy.} But we analyse the micro architecture of the SlowDepth and find out that the 2D point-wise convolution $1\times1$ serves the same purpose of the 3D $1\times1\times1$ kernel in the second stage as shown in Fig.~\ref{fig:FSB}. So we suppose that, not only it is redundant and can be removed, but also may it make the network deeper and optimization searching space wider, which hinder itself from converging to an optimal local minimum. Hence, we design another variant structure named \textbf{\emph{SlowFSB}} by resecting the redundant point-wise convolution and merging the duty of reshaping the output channel to the 3D $1\times1\times1$ kernel at the final step. From the evaluation result, it can be found that, the SlowFSB model is superior to the baseline, especially with respect to the loss and top-1 accuracy of training/validation. It is indicated that our structure is more powerful in fine-grained feature learning, considering that the improvement on top-1 accuracy needs more precise features to distinguish the subtle details. For the partial separated method in SlowFast network, the first stage employ a $3\times1\times1$ kernel to convolve data along the time direction, so each channel in the ofmap sent to the next step already contains the temporal fusion information. On the other hand, $1\times3\times3$ kernels at the second stage should be focused on learning the spatial features of image instances. Merging data from different channels here will increase the intrinsic learning workload of the spatial kernels, and disarrange the proper fusion of the spatio-temporal information with $1\times1\times1$ kernels at the final step. Hence, the proposed fully separable structure can effectively learn the spatial and temporal features step by step without disruptions, thus leading to a higher accuracy on top-1 task (training(top1/top5): from $84.1\%/97.8\%$ to $87.7\%/98.5\%$; validation(top1/top5): from $77.6\%/91.8\%$ to $79.4\%/91.3\%$). It even achieves a more compact model with a 2.3$\times$ compression rate. So we employ this variant prototype as the final FSB structure.

\begin{table}[t]
\begin{center}
\begin{tabular}{l||c|c|c}
  \hline
  Model         & Pre-train   &  Accuracy  & Model Size \\
  \hline\hline
  C3D~\cite{tran2015learning}            & from scratch        &  $47.9\%$  & 27.7M \\
  \hline\hline
  FSB-C3D           & from scratch        &  $47.5\%$  & \textbf{3.7M} \\
  \hline
  MiCT~\cite{zhou2018mict}           & from scratch        &  $56.5\%$  & 19.1M \\
  \hline
  MiCT           & ImageNet            &  $85.1\%$  & 19.1M \\
  \hline
  SlowFast~\cite{feichtenhofer2018slowfast}       & from scratch        &  $91.5\%$  & 22.46M \\
  \hline
  FastTRG           & from scratch        &  \textbf{94.1}\bm{$\%$}  & \textbf{9.9M} \\
  \hline
\end{tabular}
\end{center}
\caption{\textbf{Comparisons of model size and accuracy between the proposed FastTRG and other baselines on UCF-101.} The accuracy data of C3D and MiCT refer to~\cite{zhou2018mict}}\label{tab:accuracy}
\end{table}

\textbf{FastTRG for fast and robust learning.} Since we have significantly compressed the model size and even achieved a better classification accuracy by introducing the FSB structure to the slow path, it is wondering that if the fast path can be further improved or not. From the experiment results, it is observed that utilizing temporal convolutions in the earlier layers leads to an accuracy degradation, which is argued to be caused by those action cases that moves fast. We suppose it is because in these cases the correlation between adjacent frames is too small, so it makes the fast path harder to learned the action motions precisely. Inspired by the HOG feature~\cite{llorca2013vehicle} extractor widely used in 2D image object detection tasks, which forms the features by computing the spatial gradients of local field, we propose a heuristic method called TRG, shown in Fig.~\ref{fig:trg}, to extract temporal motion features. Intuitively, the gradient between adjacent image frames indicates the motions of the objects. We analyse the feature maps from one middle layer of the fast path, and find out that the intrinsic concept of TRG is still convincible in the early layer. We evaluate the variant, named as \textbf{\emph{FastTRG}}, which employs the TRG in the fast path and exploits the slow path of the SlowFSB. From the experiment results in Fig.~\ref{fig:design}, it can be concluded that the FastTRG model shows overall improvements over the SlowFSB and far more superiorities than the baseline, no matter in training or validation. According to the validation curves of loss, top-1 and top-5 accuracy, the TRG significantly improves the performance, which means it has better generalization ability and more robustness, in other words, it learns features closer to the real distribution of data in the video tasks. Moreover, both in training and validation phases, the FastTRG converges faster. In the early stage of learning process, it almost reaches a 10\% accuracy gap over the baseline at the same epoch. Meanwhile, considering that the fast path only costs negligible computation operations ($<3\%$), it can be concluded that the proposed TRG offers more robustness and faster convergency at little cost. We also compared the proposed FastTRG with other popular models, \wrt their model size and accuracy,  as shown in Table.~\ref{tab:accuracy}.

Although the TRG method is based on the priori acknowledge of the video tasks and forces the model to learn some heuristic manual-designed features, it unveils some intrinsic concepts of how the video-aimed 3D ConvNet actually learns and works. It also provides a instructive guideline for future works on visual tasks that a manual feature method can be applied to the end-to-end ConvNet approach and used to improve its performance. On the other hand, from the view of model compression, the TRG will introduce more sparsity to the dataflow, which can be leveraged for accelerating the inferences of video applications by sparse-dedicated hardware accelerator.



\textbf{Other variants.} There are two more variants for comparisons. The first one, noted as \textbf{\emph{Fast3D}}, shows slightly low accuracy than the baseline. It replaces the residual bottleneck structure in the fast path with traditional 3D $3\times3\times3$ kernels. So it is concluded that simply increasing the model capacity of the fast path can not improve the model accuracy, on the contrary, it proves that the TRG method in the FastTRG variant actually helps the model to learn some useful features. As for the second variant, named as \textbf{\emph{FastDepth}}, it only applies the FSB structure to the fast path. It can be seen that the compressed fast path results in considerable accuracy loss, despite of its incredibly low proportion of the entire model ($<2\%$). It demonstrates that the fast path plays an important role in the final model accuracy and a little wrong modification on its structure may cause great accuracy loss. On the other hand, the curve further proves the efficiency of the TRG method, and shows the correctness of the efforts we made on compressing the slow path, which indeed has sufficient redundancy and low correlation with the final accuracy.

\subsection{Validation on C3D Network}
\begin{figure}[h]
  \centering
  \includegraphics[width=3in]{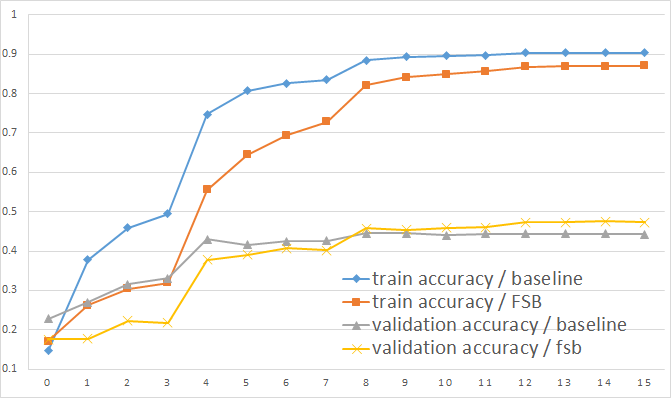}\\
  \caption{Verify the proposed methods on C3D.}\label{fig:c3d}
\end{figure}

We also evaluate the C3D ConvNet modified by the proposed compression methods on the UCF-101 to prove their adaptivity. The architecture is modified by replacing all $3\times 3\times 3$ convolution kernels with our FSB structures, as shown in Table.~\ref{tab:fsbc3d}. Because the C3D network is simply constructed by stacking convolution layers, in which batch normalization layers are not utilized. However, intensive residual operations in the TRG will result in significant network degradation when coupled with the ReLU function, \ie gradient vanishing problem. So we only employ the TRG modules in the first two layers, without batch normalization operation between blocks for fair comparison. We denote the model modified with our methods as \textbf{\emph{FSB-C3D}}. The evaluation results illustrated in Fig.~\ref{fig:c3d} shows that the FSB-C3D reaches a better validation accuracy as well as $5\times$ compression rate on convolutional layers. We argue that the performance gap between the training curves results from the severe overfitting of the C3D baseline, on the contrary, the FSB-C3D imposes strong restriction onto the architecture, forcing the model to learn intrinsic data distribution of video tasks instead of that of the training set.


\begin{table}[t]
\begin{center}
\begin{tabular}{l|c|c|c}
  \hline
  Type            & FSB-C3D                         & C3D~\cite{tran2015learning}                         & Rate \\
  \hline\hline
  $conv_1$           & FSB$_{group = 64}$          & $[64,3]$                        & $1.0\times$ \\
  \cline{1-1}\cline{3-4}
  $conv_2$           & +TRG                           & $[128,64]$                   & $10.5\times$ \\
  \hline
  $conv_3$           & FSB$_{group = 128}$              & $[256,128]$                & $10.7\times$ \\
  \cline{1-1}\cline{3-4}
  $conv_4$           & FSB$_{group = 256}$              & $[256,256]$                & $13.3\times$ \\
  \cline{1-1}\cline{3-4}
  $conv_5$           & FSB$_{group = 256}$              & $[512,256]$                & $10.7\times$ \\
  \cline{1-1}\cline{3-4}
  $conv_6$           & FSB$_{group = 512}$             & $[512,512]{}\times3$        & $6.7\times$ \\
  \hline\hline
  Total              & 27.7M                      & \textbf{3.7M}                         & \textbf{7.4$\times$} \\
  \hline
\end{tabular}
\end{center}
\caption{Architecture of FSB-C3D and compression rate of each layer over the C3D baseline. }\label{tab:fsbc3d}
\end{table}


\section{Conclusion}
In this work, we first propose a light-weight convolutional block called FSB to efficiently extract the spatio-temporal features, which can be employed to construct a extremely compact 3D ConvNet for video tasks. Evaluations on the C3D model demonstrates the effectiveness of the FSB in significantly compressing the model size by $5\times$ with even better validation accuracy, which indicates the superior modeling capacity of the proposed FSB. Furthermore, inspired by the HOG method in image feature extraction, we propose a temporal feature enhancement approach termed TRG, which remarkably improves the model performance. It is shown that the TRG offers higher accuracy, faster convergency and better robustness, while only costs negligible addition and shifting operations, which are hardware-efficient. On the other hand, we also propose a computation complexity reduction method named hFA to cooperate with the FSB for further decreasing the multiplications in a 3D ConvNet. Based on these comprehensive optimization methods, we design a light-weight network for video tasks, which shows overwhelming performance above the state-of-the-art networks. Our work indicates that there is still considerable redundancy even in the current compact-designed models, meanwhile, it also illuminates that heuristic methods like traditional feature engineering can be embedded in the popular end-to-end ConvNet models and improve the performance on specific task.


{\small
\bibliographystyle{ieee}
\bibliography{bibfile}
}

\end{document}